\documentclass[runningheads]{llncs}
\usepackage{graphicx}
\usepackage[caption=false]{subfig}
\usepackage{amssymb}
\usepackage{amsmath}

\DeclareMathOperator*{\argmin}{arg\,min}

\def\resnet{ResNet-18~}
\def\resnetv{ResNet-34~}
\def\resnetvv{ResNet-50~}
\def\dropout{ResNet-Dropout~}
\def\disturbed{ResNet-DisturbLabel~}
\def\reimp{ResNet~}

\begin{document}
\title{Phantom Embeddings: Using Embedding Space for Model Regularization in Deep Neural Networks}
\titlerunning{Phantom Embeddings}
% If the paper title is too long for the running head, you can set
% an abbreviated paper title here
%
\author{Mofassir ul Islam Arif\inst{1} \and
Mohsan Jameel\inst{1} \and
Josif Grabocka\inst{2} \and
Lars Schmidt-Thieme\inst{1}}
\authorrunning{M. Arif et al.}
% First names are abbreviated in the running head.
% If there are more than two authors, 'et al.' is used.
%
\institute{Information Systems and Machine Learning Lab, University of Hildesheim, Germany  \\
            \email{$\{$mofassir,mohsan.jameel,schmidt-thieme$\}$@ismll.uni-hildesheim.de}
        \and Department for Computer Science,Albert-Ludwigs-University, Freiburg,Germany\\
            \email{$\{$grabocka$\}$@informatik.uni-freiburg.de}}
\maketitle              % typeset the header of the contribution
\begin{abstract}
The strength of machine learning models stems from their ability to learn complex function approximations from data; however, this strength also makes training deep neural networks challenging. Notably, the complex models tend to memorize the training data, which results in poor regularization performance on test data. The regularization techniques such as L1, L2, dropout, etc. are proposed to reduce the overfitting effect; however, they bring in additional hyperparameters tuning complexity. 
These methods also fall short when the inter-class similarity is high due to the underlying data distribution, leading to a less accurate model.

In this paper, we present a novel approach to regularize the models by leveraging the information-rich latent embeddings and their high intraclass correlation. We create phantom embeddings from a subset of homogenous samples and use these phantom embeddings to decrease the inter-class similarity of instances in their latent embedding space. The resulting models generalize better as a combination of their embedding, regularizes them without requiring an expensive hyperparameter search. We evaluate our method on two popular and challenging image classification datasets (CIFAR and FashionMNIST) and show how our approach outperforms the standard baselines while displaying better training behavior.

% In this paper, we present a novel approach to regularize the models by using the information rich embedding space directly to increase the overall accuracy while at the same time avoiding increasing the number of hyper-parameters to train. Our approach outperforms the vanilla image classification models in their final accuracies and also show better training behavior. We present our results on the CIFAR, and FashionMNIST datasets.

\end{abstract}
%
%
%
%!TEX root = ../main.tex
\section{Introduction}\label{sec:intro}

%===== Mohsan added ==========
The field of computer vision has seen a remarkable increase in capability and complexity in recent years. The use of deep learning models in image classification \cite{krizhevsky2012imagenet} and object detection \cite{girshick2015fast} tasks have shown a marked increase in their ability to capture more complex scenarios. Increasingly complex deep learning models such as ResNet \cite{he2016deep} and Inception \cite{szegedy2015going} were able to capture more in-depth information from input data. The strength of these deep learning models comes from their ability to take complex data and reduces it to highly expressive latent representations. These latent representations encode an image's spatial information into a vector through repeated convolutions and pooling operations.

Training these complex models bring their challenges. Generally, the true distribution of the data is unknown, and observations are available in a limited number. These models are trained by iteratively minimizing the empirical risk over the training data (also known as Empirical Risk minimization ERM \cite{vapnik1998statistical}). However, the increasing complexity of the model tends to overfit the data and generalize poorly on the test data, despite using the proper regularization. The theoretical understanding of ERM guarantees convergence as long as the model complexity does not increase with the number of training data \cite{vapnik2015uniform}. For deep neural networks, an obvious issue arises as the increase in model complexity is not always complemented by an increase in the training data.

\begin{figure}[]
\centering
\subfloat[Training Data\label{fig:toy1}]{%
  \includegraphics[width=0.33\textwidth]{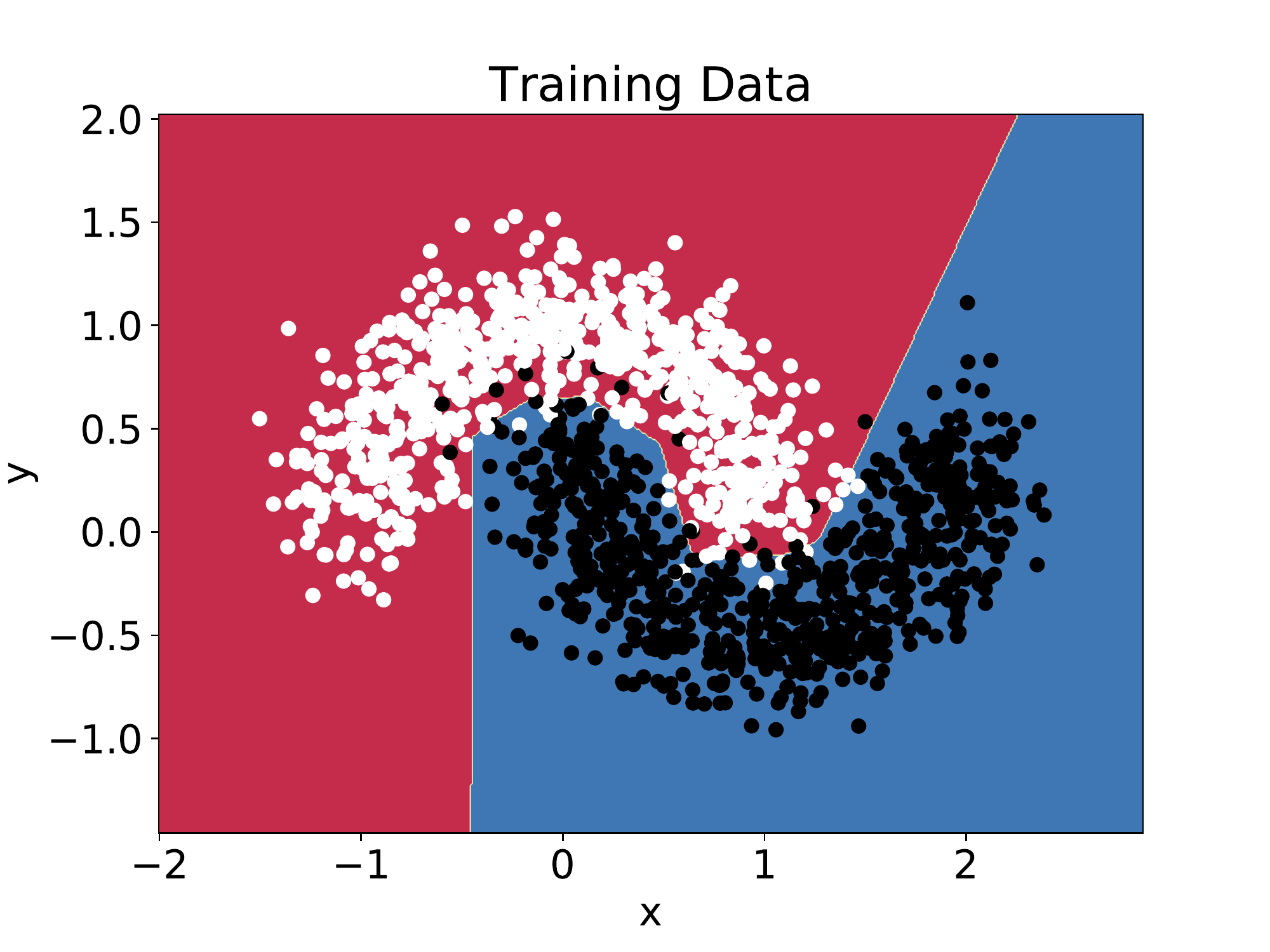}%
}\hfil
\subfloat[Testing Data\label{fig:toy2}]{%
  \includegraphics[width=0.33\textwidth]{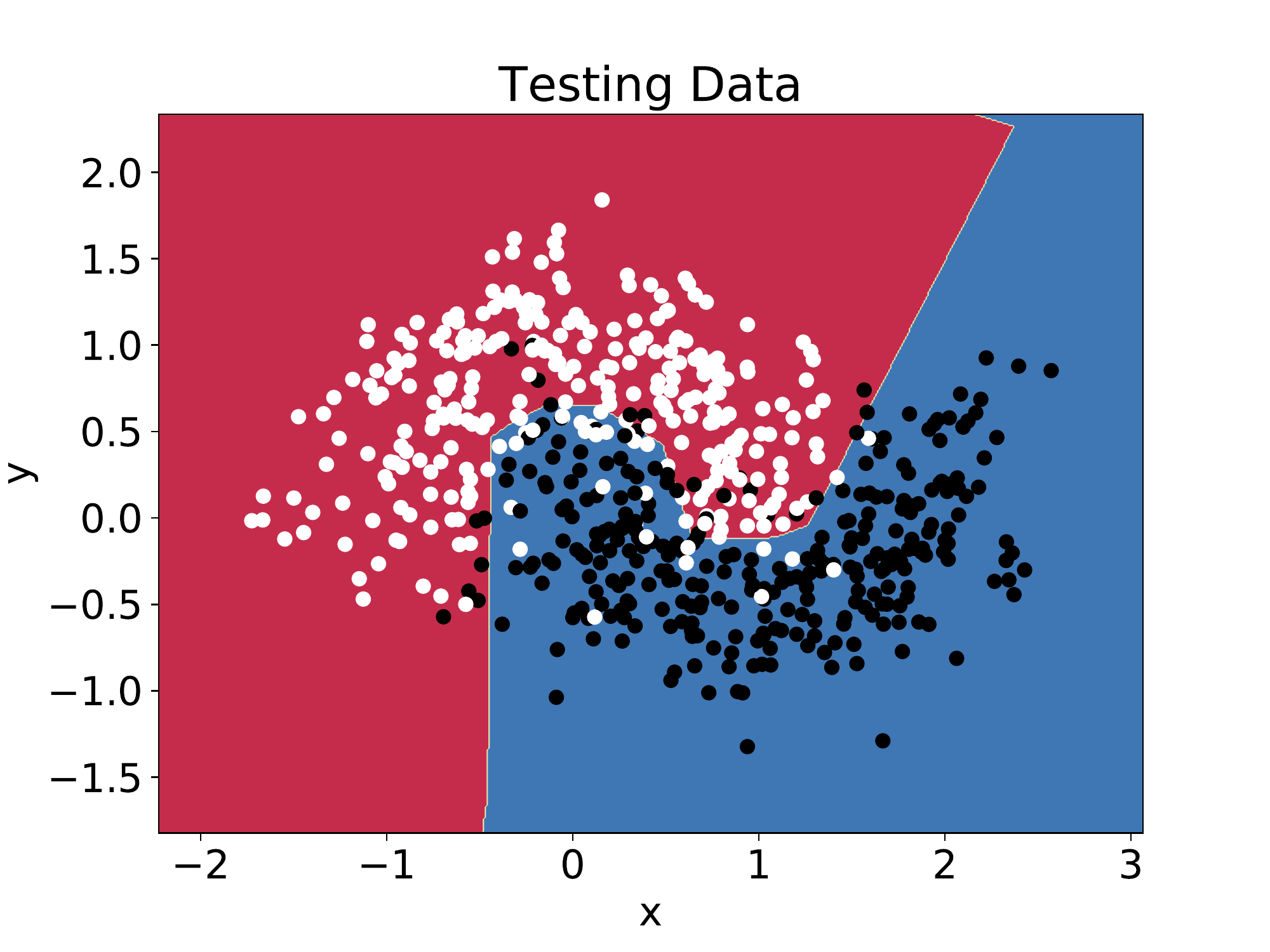}%
}
\subfloat[Phantom Embeddings\label{fig:toy3}]{%
  \includegraphics[width=0.23\textwidth]{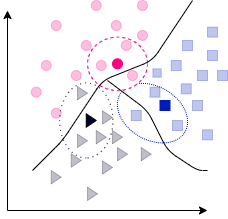}%
}
\caption{ Fig. \ref{fig:toy1} shows the overfitted decision boundary on training data. When evaluated on the test set in Fig. \ref{fig:toy2} the model shows poor generalization, a consequence of overfitting. In Fig. \ref{fig:toy3}, we show a hypotethical embedding space and the decision boundary created by a deep neural network. The light colors represent the original embeddings while the darker colors represent the phantom embeddings proposed by our method. }
\label{fig:toy}
\end{figure}
%===== Mohsan added ==========
To illustrate the aforementioned problems, we train a feed-forward neural network on a synthetic binary classification dataset and visualize the decision boundary in Fig. \ref{fig:toy}. Fig. \ref{fig:toy1} shows that the model was able to learn a reasonable decision boundary on the training data. However, due to the limited training examples available to train a complex model, it could not capture a better generalizable decision boundary resulting in poor performance on the test examples, as shown in Fig. \ref{fig:toy2}.
This example showcases two crucial challenges, firstly, how easy it is to overfit and perform poorly on test data. Secondly, in Fig.1a it can be seen that certain instances from differing classes are very close to each other, and ERM fails to provide a procedure to capture those instances.

The model overfitting is treated by introducing the regularization \cite{goodfellow2013maxout,krogh1992simple,srivastava2014dropout,ioffe2015batch} in the ERM objective. However, ERM's problem is most evident around the vicinity of the boundary region, as samples from different classes are in close proximity. Model complexity could be increased to capture these instances, but that violates the convergence guarantee of ERM since the number of instances does not increase with the increase in model complexity. One can mitigate the ERM failure through the Vicinal Risk Minimization Principle \cite{chapelle2001vicinal} by adding a better regularization using data augmentation \cite{simard1998transformation}. Data augmentation mutates the input instances, traditionally through rotating, flipping, and scaling to inject noise in the training data, thereby preventing the model from memorizing it. However, it is limited as it mutates the data within one class vicinity and not across other classes. Other regularization methods involve tunable hyperparameters requiring an expensive configuration search, and the resulting hyperparameters are non-transferable and dataset-specific.
In this paper, we propose a solution for problems stated above by leveraging the latent embeddings to create what we call a 'phantom embedding'. This is done by aggregating the latent embeddings of a subset of the instances from the same class. Using the latent vicinal embedding space allows us to use the information-rich embeddings to inject a hyper-parameter free latent vicinal regularization and boost accuracy. Since machine learning models transfrom input data into their representative embeddings: $\psi: \mathbb{R}^M \rightarrow \mathbb{R}^D$ where $M$ is the original data dimensionality and $D$ is the size of the embedding space. Therefore, by creating this phantom embedding, we create phantom data points to learn on. This is illustrated in in Fig. \ref{fig:toy3}.
This phantom embedding is used to `pull' the original instance away from the decision boundary and closer to the samples (of the same class) in the embedding space. For the instances already sufficiently away from the decision boundary the `pull' does not adversely impact since the embedding space is already well seated in the data distribution. We validate on an image classification benchmark task that our propose solution generalize better as compare to the existing approaches and achieves higher test accuracies.

% We describe our proposed method in Section \ref{sec:method} and then proceed to show our experimental findings in Section \ref{sec:results} where we compare our method's performance on the CIFAR\cite{krizhevsky2014cifar} and FashionMNIST \cite{xiao2017fashion} datasets . Our main contributions include:

Our main contributions include:
\begin{itemize}
\item Improvement in classification accuracy by using phantom data points to overcome the base error in a dataset.
\item A hyper-parameter free intrinsic regularization to enable training truly deep models.
\item Evaluate our model on two popular datasets against established baselines and showcase our performance gains as well training improvement qualitative and quantitatively.
\end{itemize}

%  This is show in Eq. \ref{eq:comb}

% \begin{equation}
% \psi(x) = \lambda\psi(x_i) + (1-\lambda)\psi(x_j)
% \label{eq:comb}
% \end{equation}

 %!TEX root = ../main.tex
\section{Related Work}

Training very deep networks effectively is an open question\cite{srivastava2015training} due to the model complexity. Models with millions of parameters require a lot of data to train effectively, however, millions of training samples are not available for all tasks. A good example of the realistic amount of data needed is \cite{deng2009imagenet} with 16M instances. That is not an option for all machine learning settings especially domains such as medicine \cite{sirinukunwattana2016locality}. Data augmentation \cite{krizhevsky2012imagenet} is an efficient method to ensure that data seen by the model is varied during training. Standard augmentation techniques include flipping, scaling, and padding.

 Training these models from scratch can be avoided by using the weights of a model that has been trained on a similar dataset and then finetuning the model to fit your need\cite{pratt1993discriminability} \cite{simonyan2014very}. Transfer learning \cite{pratt1993discriminability} has enabled training deeper model using a smaller dataset size however, when the goal is complete retraining than the training procedure needs to be adapted to ensure that the model does not memorize the training data.

Methods such as MaxOut \cite{goodfellow2013maxout} add layers into the architecture with a max activation function and have shown to positively impact the convergence behavior when compared to the ReLu activation \cite{nair2010rectified}. DropOut, proposed in \cite{srivastava2014dropout}, addresses the problem of model overfitting by probabilistically turning off neurons in the final embedding layer to create an ensemble of models and has shown to be an effective way to regularize deep neural networks. Similarly in \cite{xie2016disturblabel}, the authors move the regularization from the final layer to the loss layer where they intentionally flip the labels in a mini-batch to ensure that the model generalizes. These methods seek to work on the architecture and loss layer to regularize the model. Methods such as weight decay \cite{krogh1992simple} and batch normalization\cite{ioffe2015batch} are aimed at the optimizer and architecture and seek to penalize the weights while training to ensure that models generalize.

In \cite{zhang2017mixup} the authors propose the use of taking multiple instances and creating a linear combination of the instances and their label. Sampling from this mixup distribution allows them to learn on fabricated data points.

% https://emerj.com/ai-sector-overviews/machine-vision-in-insurance-current-applications/

%!TEX root = ../main.tex
\section{Methodology}\label{sec:method}

\begin{figure}
\centering
  \includegraphics[width=0.8\linewidth]{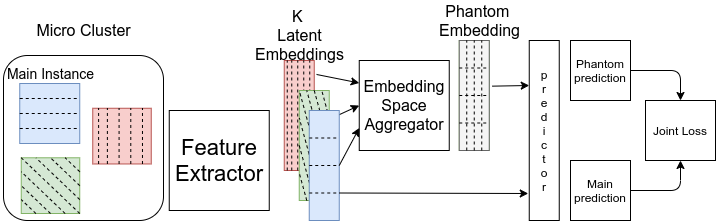}
  \caption{A training steps takes a micro-cluster with K many samples, generating K many embeddings with are aggregated to create the phantom embedding, This, along with the embedding of the main train instance is passed to the predictor. The combined loss for these predictions is calculated as in Eq. \ref{eq:loss}.}
  \label{fig:design}
\end{figure}

Consider a machine learning method $\psi(x)$ where $x$ is a dataset sample and $x \in \mathbb{R}^{N\times M}$ corresponds to a multi-category target $y$ where $y \in \{ 1,\dots,L\}^N$ among $L$ classes. This model will produce a latent embedding: $\phi : \mathbb{R}^M \rightarrow \mathbb{R}^D$ of the features, which is then passed to the prediction layer: $\psi: \mathbb{R}^D \rightarrow \mathbb{R}^L$. The estimated target variable is therefore $\hat{y_n}:= \psi(\phi(x_n)) , \forall n \in  \{ 1,\dots,N\}$ and the respective objective function:

\begin{equation}
\argmin _{\psi,\phi} \sum_{n=1}^{N} \mathcal{L}(y_n , \psi(\phi(x_n)))
\label{eq:bb_formal}
\end{equation}
In this work we propose to make use of the shared similiarities among the instances belonging to the same class and leveraging the collective learned representations of a small subset of instances to generalize the final embedding space. This is done by sampling a `micro-cluster' of instances belonging to the same class. Note here that `cluster' is being used in terms of a `group' and has no relation to the unsupervised clustering methods.

Let us denote the number of instances in each micro-cluster as $ K \in \mathbb{N} $and
the number of instances in each respective class as $N_l \in \mathbb{N}, \forall l \in \{1,\dots,L\}$ therefore for each class it is possible to draw $N_l \choose K$ many random choices. On these choices, consider, a new dataset transformation $(x, y) \rightarrow (x' , y')$, where each
element of $x'$ represents a homogeneous cluster from $x$ with $K$ members and each
element $y'$ is the respective label of the instances within a homogeneous
cluster. Since we are sampling homogenous clusters, $y'=y$. The total number of clusters is defined as $N' = \sum_{l=1}^{L} {N_l \choose K}$. The new input features are then  $x' \in \mathbb{R}^{N' x K x M}$ and the new targets $y' \in \{1,\dots,L\}^{N'}$. 

This new dataset transformation leads to a model output: $\hat{y_n}:= \psi(\phi(x'_{n,k}))$ where $\phi(x'_{n,k})$ is the $k^{th}$ latent embedding and $k \in K$. These $K$ many latent embeddings will be used to generalize the learned final learned embedding by aggregating them as see in Fig. \ref{fig:design}. In our proposed approach we use a "Mean Emdedding Space Aggregator" which is explained as:
$\phi'(x_n) = \frac{1}{K}\sum_{k=1}^{K}\phi(x'_{n,k})$ where $\phi'(x_n)$ is the phantom embedding from the micro-cluster. The naive approach would be to use this phantom embedding  directly in the optimization,resulting in the folllowing objective function:

\begin{equation}
\argmin _{\phi,\psi} \sum_{n=1}^{N'} \mathcal{L}\bigg(y'_n , \psi \big(\frac{1}{K} \sum_{k=1}^{K}\phi(x'_{n,k})\big)\bigg)
\label{eq:naive}
\end{equation}
However, Eq. \ref{eq:naive} poses a problem since the intra-class variation of challenging datasets can cause the embedding to be too drastically modified, Also, datasets with multi-modal distributions and non-convex hulls can be adversely effected by the naive objective function (Eq. \ref{eq:naive}) since the micro-cluster can be sampled from the different modes of the data distribution. 
In its place we propose to use the phantom embedding in the loss function:
% to guide the embedding spaces away from train dataset memorization. Our proposed loss is:

\begin{equation}
\mathcal{L} = \alpha \mathcal{L}(y'_n , \psi(\phi(x'_{n,k=0}))) -  (1-\alpha)\mathcal{L}(y'_n , \psi(\phi'(x'_{n})))
\label{eq:loss}
\end{equation}
In Eq. \ref{eq:loss} we treat the first sample ($k=0$) as the main instance and the others serve as a guide to improve the embedding space for this instance by `pulling' the $k=0^{th}$ towards the phantom embedding. In order to avoid adding another hyper-parameter to tune, in our loss, $\alpha$ is drawn from the beta distribution and serves to add stochasticity in the combination of the embeddings and also removes the need for tuning  $\alpha$. Therefore our final objective function is:
\begin{equation}
\argmin _{\phi,\psi} \sum_{n=1}^{N'} \bigg[ \alpha \mathcal{L}\bigg(y'_n , \psi\big(\phi(x'_{n,k=0})\big)\bigg) -  (1-\alpha) \mathcal{L}\bigg(y'_n , \psi \big(\frac{1}{K} \sum_{k=1}^{K}\phi(x'_{n,k})\big)\bigg) \bigg]
\label{eq:obj}
\end{equation}

%!TEX root = ../main.tex

\section{Experiments}\label{sec:results}

In this section, we showcase the results of our approach and compare them with other methods in the domain.
All the results presented have been recreated using the original author's provided implementations.
These experiments were carried out on NVIDIA 1080Ti, 2080Ti, and V100 GPUs.
\subsection{Datasets and Implementation Details}

To verify the efficacy of our proposed approach we have chosen two publically available datasets. CIFAR10 \cite{krizhevsky2014cifar} and FashionMNIST \cite{xiao2017fashion} are popular image classification datasets and are widely used in the computer vision domain for testing new research. They comprise 60000 and 70000 images sized at 32x32 and 28x28 respectively. They offer a challenging problem setting due to the wide intra-class variation and inter-class similarities. Furthermore, these datasets are also easy to overfit the deep convolutional neural networks. Therefore, these datasets provide all the necessary challenges that our work proposes to address.

Our method can be readily included in any machine learning model, for our experiments we have chosen Deep Residual Networks (ResNet-18, ResNet-34, and ResNet-50) as proposed in \cite{he2016deep} and as implemented in  \cite{kli}. 
%One important fact to note here is that He. et al proposed their model with the ImageNet dataset as their test case. The original ResNet architecture comes with a kernel size of 7 for the first convolutional layer, however that leads to poor performance for CIFAR and similar datasets due to the shrinking receptive field. This implementation we have used uses a kernel of size 3 and has shown to be an effective choice. 
The networks under test were initialized as specified in \cite{he2015delving} and optimized using Stochastic Gradient Descent(SGD) \cite{lecun1989backpropagation} with batch normalization \cite{ioffe2015batch} and a weight decay \cite{krogh1992simple} factor of 0.0005, it should be noted here that the original ResNet architecture used 0.0001. The learning rate was set at 0.1 at the start than the scaled down by a factor of 10 at the 32k and 48k iteration as in \cite{he2016deep}, training was terminated at 64k iterations. We used a batch size of 128 and the dataset was augmented by padding 4 pixels to the image and translating the image accordingly, the images were also flipped horizontally and normalized by the mean and standard deviation of the entire dataset.
%For the parameters specific to our model,
% we found $K=2$ to be sufficient to achieve the performance gain over the baselines and $\alpha=1.0$ for CIFAR while $\alpha=0.4$ for FashionMNIST led to the best performance.  

\subsection{Results}
%Before proceeding with the results, we reiterate our initial goals with this research. 
In this section we evaluate our model by answering the following research question:.

%\textbf{RQ1: Accuracy Improvement}, highlight how our model can lead to higher accuracy, \textbf{RQ2: Robustness}, demonstrate how learning better latent representations can mitigate misclassification among highly similar classes. \textbf{RQ3: Intrisic Regularization}, how a more rich embedding structure drawn from the dataset itself can lead to regularization behavior that is not dependant on extensive hyper-parameter tuning.
\begin{enumerate}
\item \textbf{RQ1}: Can classification accuracy be improved by creating a phantom embedding for data points?
\item \textbf{RQ2}: Can a better embedding space lead to a more robust model?
\item \textbf{RQ3}: Can we add intrinsic regularization by using the embedding space direclty?
\end{enumerate}

\subsection{RQ1: Classification Accuracy}

The baselines were chosen based on their relevance to the approach that we have outlined in this paper. We have used the DisturbLabel \cite{xie2016disturblabel} as implemented in \cite{code_DL}, ResNet with Dropout \cite{srivastava2014dropout} and we also compare against the vanilla variants of the ResNet architectures. DisturbLabel seeks to regularize the loss layer rather than the parameters and DropOut seeks to create an inherent ensemble of neural networks by stochastically turning off a certain amount neurons in the embedding layer to prevent the models from learning the training data. A comparison of our method to the baselines can be seen in Tab. \ref{tab:18}.
\begin{table}[]
\centering
\caption{Classification Accuracy on CIFAR using \resnet architecture. We report the final accuracy as \textbf{Acc} and also the \textbf{Mean} and \textbf{Max} accuracies for the last 5 epochs to illustrate training stability towards convergence.}

\begin{tabular}{|l|l|l|l|}  
\hline
\multicolumn{1}{|l|}{\textbf{}}             & \multicolumn{3}{c|}{\textbf{Accuracy}}	\\ \hline				
\textbf{Method}                & \textbf{Acc}   & \textbf{Mean Acc}  & \textbf{Max}   \\ \hline
%ResNet18 \cite{he2016deep}             & 91.2  &       &       \\ \hline
ResNet18        & 93.5  & 93.68 & 93.68 \\ \hline
ResNet18 Dropout      & 94.11 & 94.09 & 94.21 \\ \hline
%ResNet18 MixUp        & 94.48 & 94.39 & 94.48 \\ \hline
ResNet18 DisturbLabel & 94.2  & 94.28 & 94.33 \\ \hline \hline
\textbf{Phantom ResNet18 }       & \textbf{94.91} & \textbf{94.84} & \textbf{94.91} \\ \hline \hline
\end{tabular}
\label{tab:18}
\end{table}

%It should be noted here that the \textbf{`ResNet18'} refers to the values as mentioned in the original paper. Upon our reimplementation and the use of better kernel size (3x3), we were able to see a lift over the original. Therefore, we will be using \textbf{`ResNet '} as our primary baseline and all other methods will be compared to it. 

% We report the final test accuracy as well as the behavior of the model leading up to convergence by looking at the mean of the last 5 epochs to show the stability of the model as in \cite{srivastava2015training}.  

It can be seen in Tab. \ref{tab:18} and \ref{tab:34} that our proposed method is performing better than the all the baselines in terms of the accuracy, however, it should also be noted that the overall variance in the results at the time of convergence is also better than the baselines.

\begin{figure}[]
\centering
\subfloat[Train loss\label{fig:resnet18_train_loss}]{%
  \includegraphics[width=0.33\textwidth]{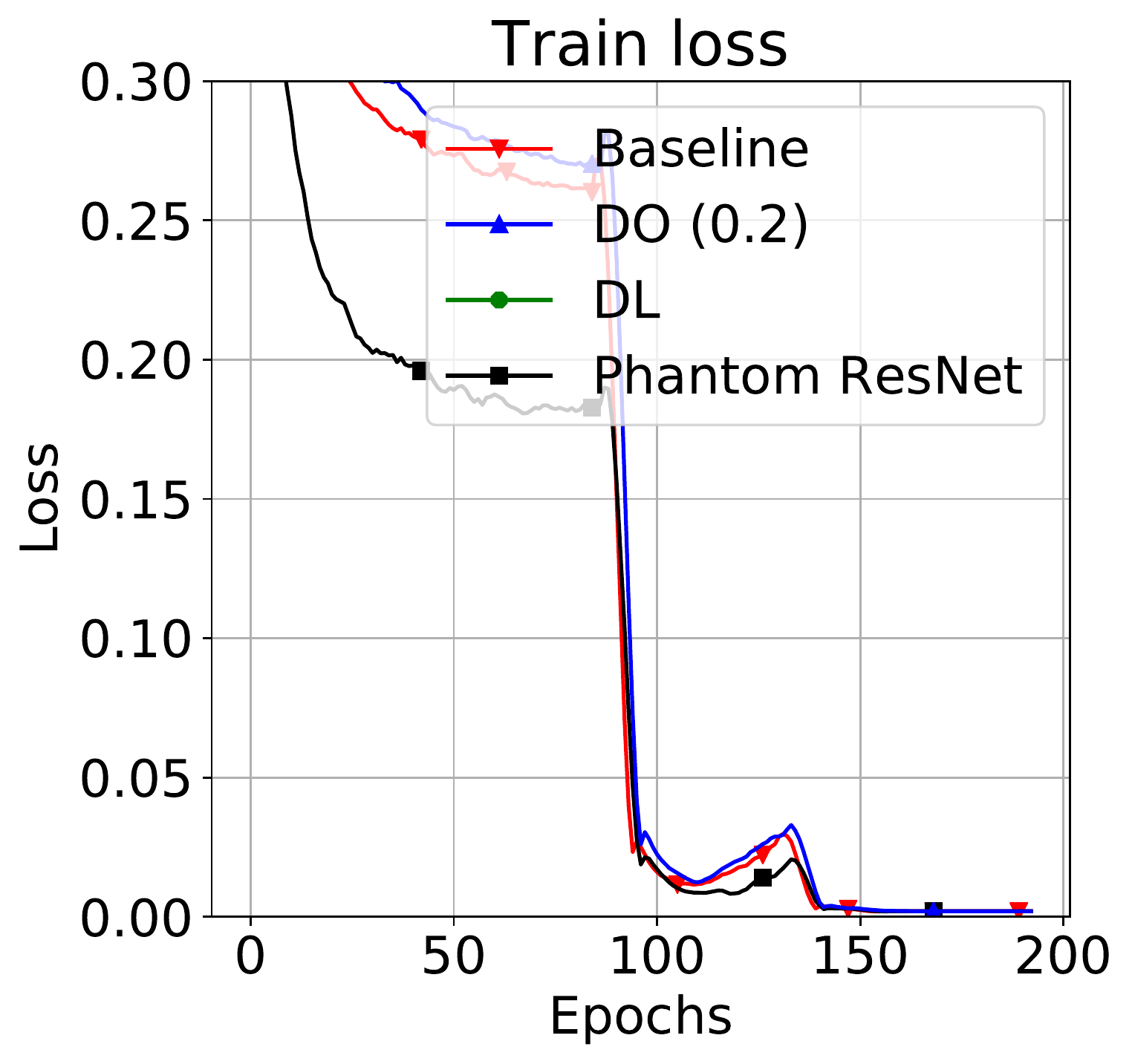}%
}\hfil
\subfloat[Test loss\label{fig:resnet18_test_loss}]{%
  \includegraphics[width=0.33\textwidth]{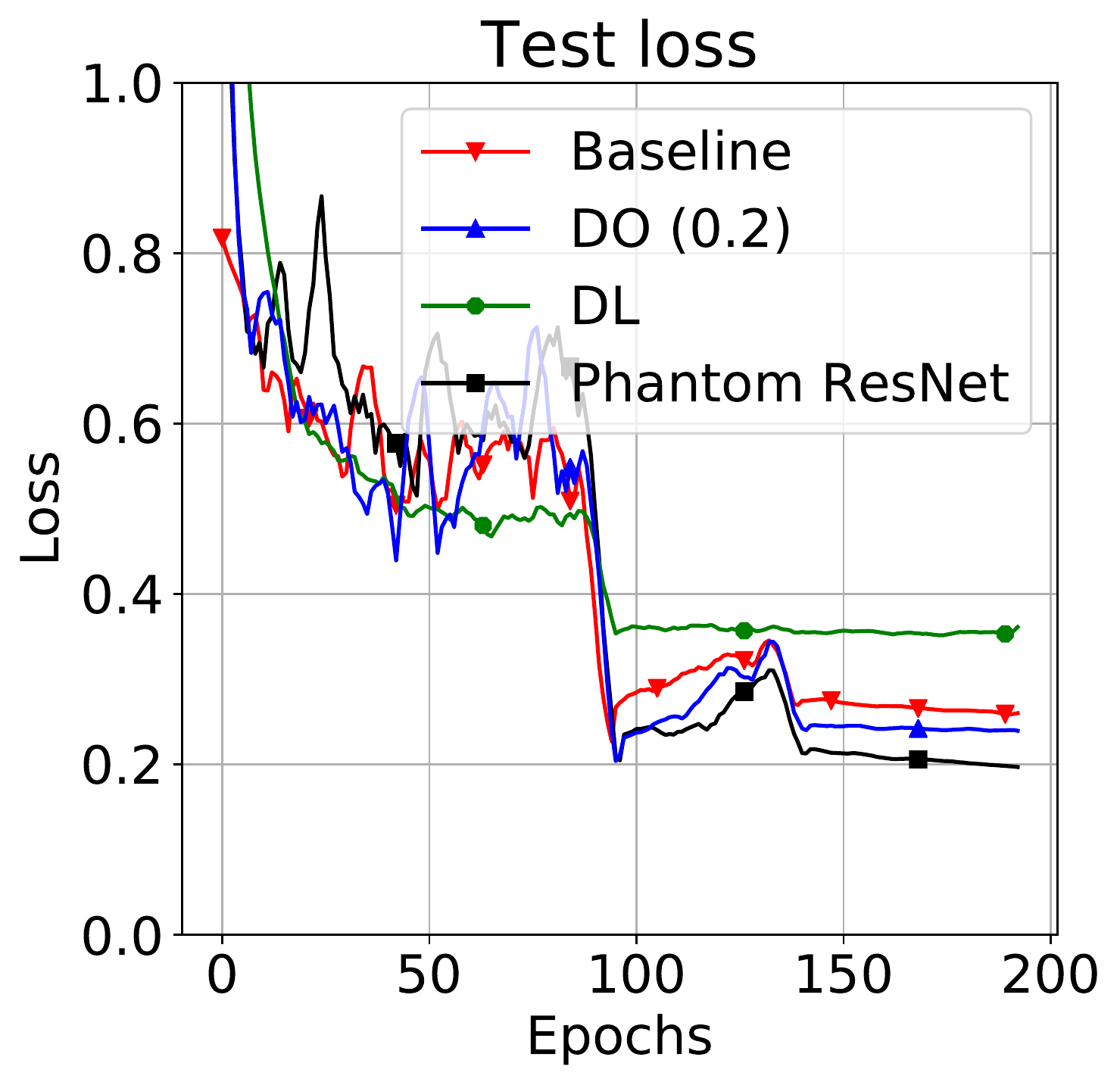}%
}\hfil
\subfloat[Test Accuracy\label{fig:resnet18_test_acc}]{%
  \includegraphics[width=0.33\textwidth]{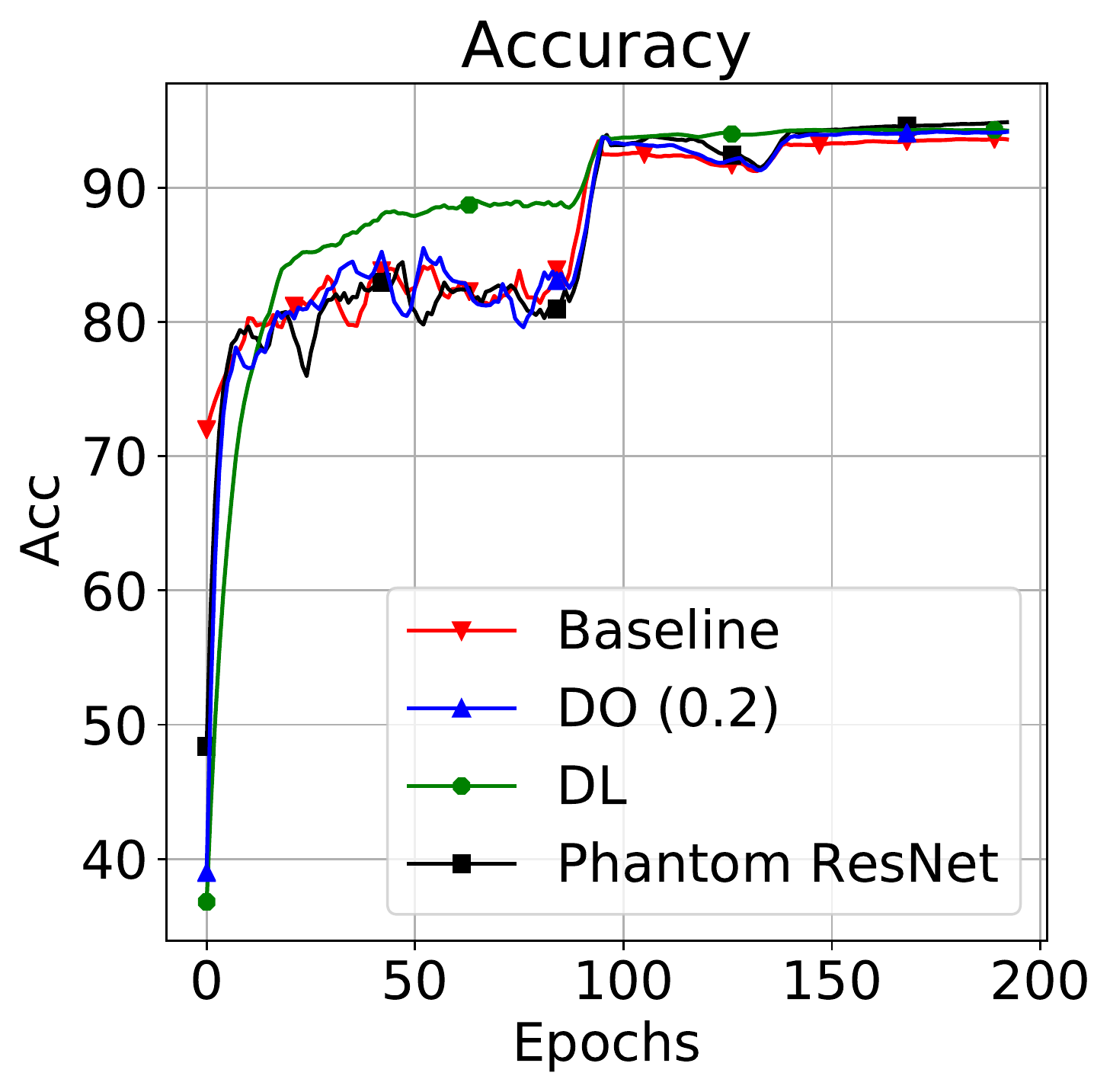}%
}

\caption{\resnet Training and testing behaviors: \textbf{Baseline} refers to the original network baseline while \textbf{DO} and \textbf{DL} refer to the DropOut and DisturbLabel baselines.}
\label{fig:resnet18_plots}
\end{figure}

\begin{table}[]
\centering
\caption{Classification Accuracy on CIFAR using ResNet34}

\begin{tabular}{|l|l|l|l|}
\hline
\multicolumn{1}{|l|}{\textbf{}}             & \multicolumn{3}{c|}{\textbf{Accuracy}}	\\ \hline	
\textbf{Method}            & \textbf{Acc}   & \textbf{Mean}  & \textbf{Max}   \\ \hline
%ResNet34\cite{he2016deep}                & 92.49 &    -   &    -   \\ \hline
ResNet34           & 93.65 & 93.71 & 93.79 \\ \hline
ResNet34 Dropout       & 93.92 & 93.97 & 94.03 \\ \hline
%\textbf{ResNet34 MixUp}         & 95.72 &       &       \\ \hline
ResNet34 DisturbLabel   & 93.73 & 93.79 & 93.81 \\ \hline \hline
\textbf{Phantom ResNet34}         & \textbf{94.52} & \textbf{94.52} & \textbf{94.6}  \\ \hline \hline
\end{tabular}
\label{tab:34}
\end{table}

 %Specifically, we see a $3.71\%$ and $1.59 \%$ increase over both the original \resnet and our reimplementations respectively. 
 %For \resnetv we see a similar trend however, now the gain over the original and reimplementation is $2.03 \%$ and $0.88 \%$ respectively. 
 We see a $1.59\%$ and $0.88\%$ gain for \resnet and \resnetv accuracies, the
  decrease in the overall `performance gain' can be attributed to the fact that \resnetv is a more complex model. \resnet has 0.27M parameters while \resnetv has 0.46M, so by doubling the parameters of the network we expect a more expressive model that already improves upon the shortcomings of the former. A more important trend in Tab. \ref{tab:34} is the behavior of \dropout and \disturbed values. We only see an improvement over \reimp of $0.27\%$ and $0.1\%$. The baselines struggle to keep up with the expected increase in performance expectations when increasing model complexity. ERM convergence guarantee, as stated early is at play here. Secondly, the baselines do not take into account the highly similar embeddings of dissimilar class members. Therefore, the baselines end up optimizing for performance without taking into account the obstacles that are hindering it and it gets reflected in the final numbers we present here.  
\begin{table}[]
\centering
\caption{Classification Accuracy on FashionMNIST}

\begin{tabular}{|l|l|l|l|}
\hline
\multicolumn{1}{|l|}{\textbf{}}             & \multicolumn{2}{c|}{\textbf{Accuracy}}	\\ \hline	
\textbf{Method}        & \textbf{ResNet-18} & \textbf{ResNet-34} \\ \hline% & \textbf{ResNet-50} 
ResNet          & 94.78              & 94.93              \\ \hline% & 94.02              
ResNet-Dropout        & 94.97              & 95.11              \\ \hline% &                    
ResNet-Disturb        & 94.95              & 94.97              \\ \hline \hline% &                    
\textbf{Phantom ResNet} & \textbf{95.07}              & \textbf{95.38}     \\ \hline \hline% & \textbf{94.88}     
\end{tabular}
\label{tab:fashion}
\end{table}

In Tab. \ref{tab:fashion} we can see that the results for our approach continue to out-perform the baselines on the FashionMNIST dataset which comes with its own set of challenges since the images are now 28x28 and comprise of a single channel rather than the standard RGB channels of CIFAR. 

Consistent accuracy improvement across these datasets and over varying architecture complexities shows that our method is robust enough to deal with a wide variety of scenarios. Furthermore, it should be pointed out that the accuracies for the baselines required a large hyper-parameter search to get to these values whereas our proposal required no such search for performance.

\subsection{RQ2: Robustness}

\begin{figure}[t]
\centering
\subfloat[Train loss\label{fig:resnet34_trainLoss}]{%
  \includegraphics[width=0.33\textwidth]{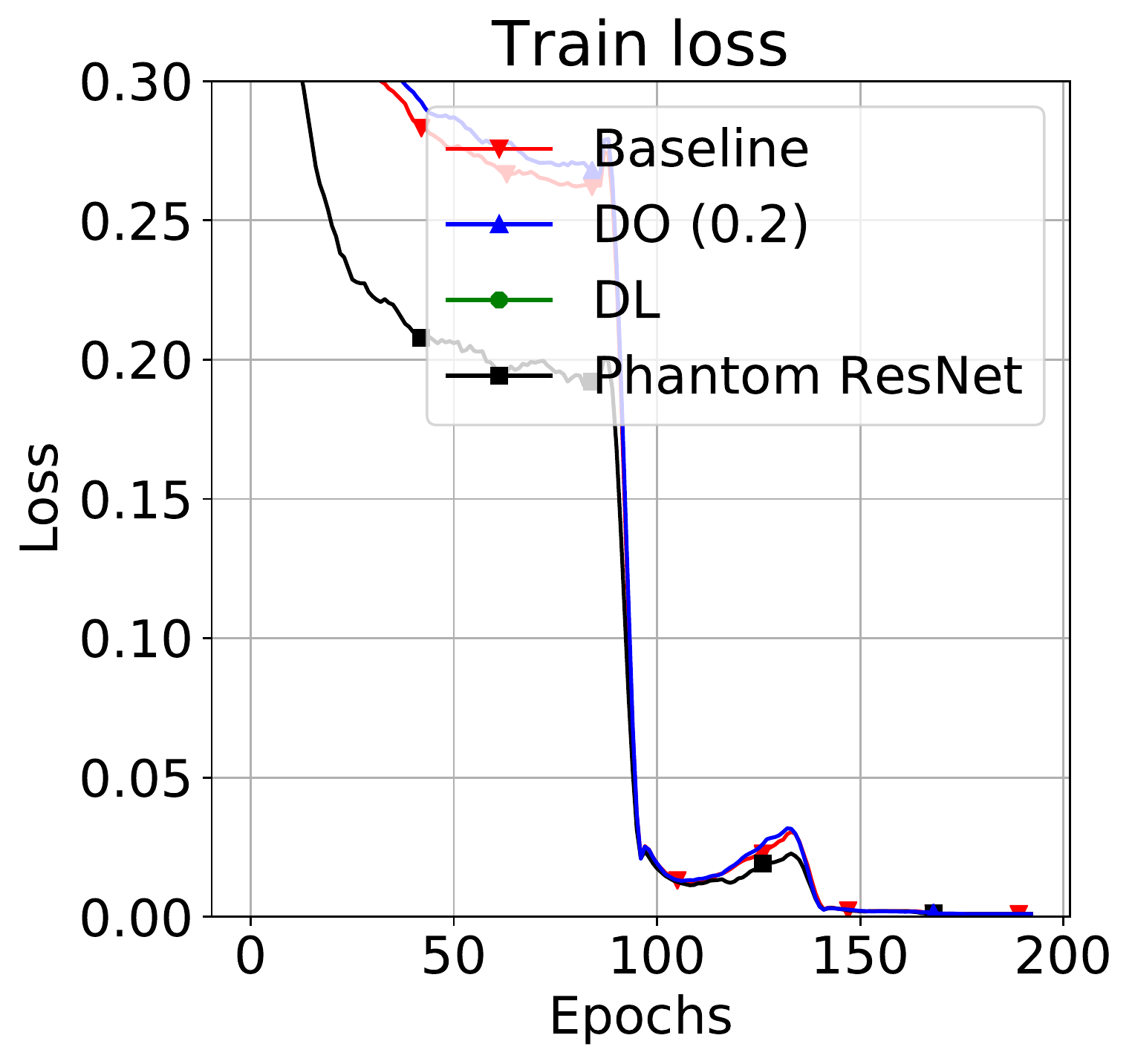}%
}
\subfloat[Test loss]{%
\label{fig:resnet34_testLoss}
  \includegraphics[width=0.33\textwidth]{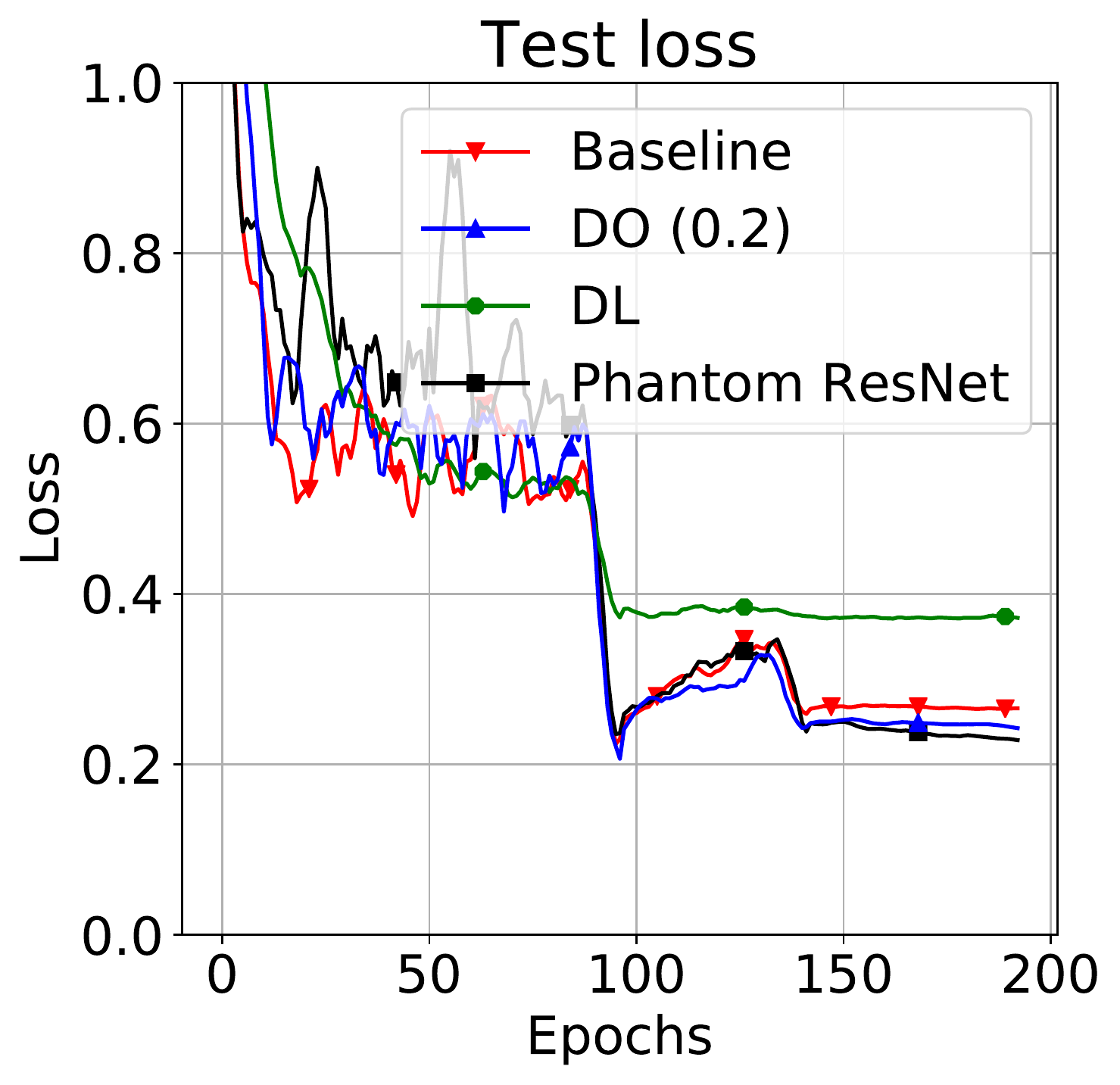}%
}\hfil
\subfloat[Test Accuracy\label{fig:resnet34_acc}]{%
  \includegraphics[width=0.33\textwidth]{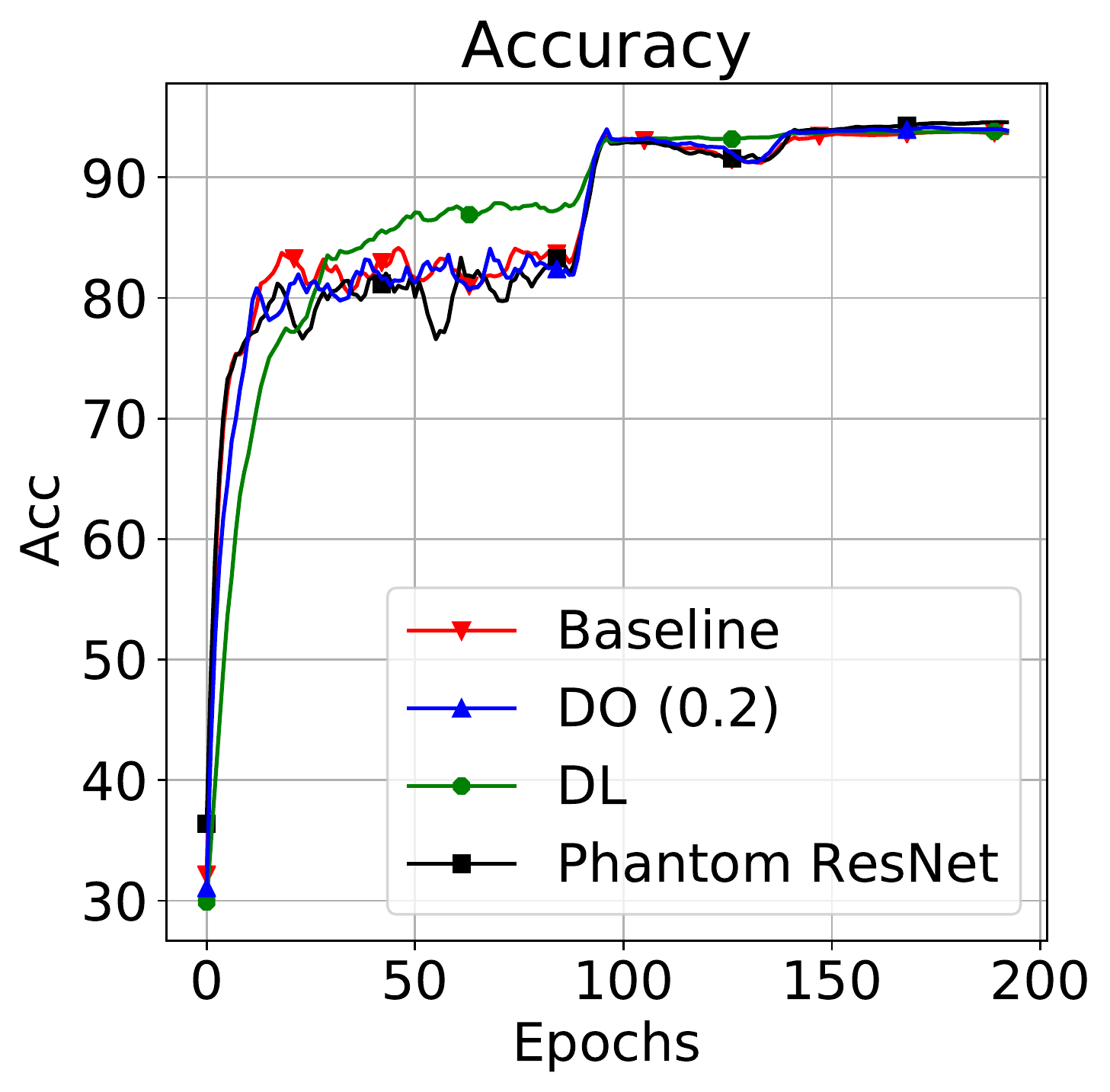}%
}

\caption{\resnetv Training and testing behaviors: 
%\textbf{Baseline} refers to the original network baseline while \textbf{DO} and \textbf{DL} refer to the DropOut and DisturbLabel baselines. 
With a more complex network, our continues to outperform the baselines.}
\label{fig:resnet34_plots}
\end{figure}
A sufficiently well-trained algorithm should be able to reduce the error on the test set, the reduction of test error is inextricably tied to the training process. 
%With the wide-scale use of deeper and deeper networks, overfitting and train data memorization is an ever-present pitfall that many models fall into. 
Our proposed methods seeks to mitigate overfitting by enriching the embedding space ensuring that the model generalizes well thus preventing errors in similar classes. 
It can be seen in Fig. \ref{fig:resnet18_test_loss} that our model is converging to a lower Test loss, this is an outcome of the enriched embedding space that actively helps optimize the model to learn a more general representation from the training data. The outcome of this approach reflects readily in Tab. \ref{tab:18} in the final accuracies, furthermore considering Fig. \ref{fig:resnet18_test_acc} it can be seen that our proposed model takes a more deliberative approach in the initial learning stage up to the first 100 epochs. While other models are shooting up quickly in accuracy values, and then later failing to maintain their lead, our approach focuses on learning better representations and penalizing itself when it doesn't more aggressively in order to arrive at the better final optimal model.

\begin{table}[]
\centering
\caption{Classification Accuracy on CIFAR using ResNet50}

\begin{tabular}{|l|l|l|l|}
\hline
\multicolumn{1}{|l|}{\textbf{}}             & \multicolumn{3}{c|}{\textbf{Accuracy}}	\\ \hline	
\textbf{Method}                          & \textbf{Acc}   & \textbf{Mean}  & \textbf{Max}   \\ \hline
%ResNet50\cite{he2016deep}              & 93.03          &                &                \\ \hline
ResNet50         & 93.86          & 93.25          & 93.34          \\ \hline
ResNet50 Dropout      & 93.21          & 93.17          & 93.25          \\ \hline
ResNet50 DisturbLabel & 94.37          & 94.352         & 94.38          \\ \hline \hline
\textbf{Phantom ResNet50}        & \textbf{94.48} & \textbf{94.54} & \textbf{94.71} \\ \hline
\end{tabular}
\label{tab:50}

\end{table}

The same trend is observed when training \resnetv as shown in Fig. \ref{fig:resnet34_plots}. The only difference being that models no trained with inherent embedding space enrichment in mind suffer more due to the higher complexity of the underlying networks.
In both Fig. \ref{fig:resnet18_plots} and Fig. \ref{fig:resnet34_plots} it can be seen that \dropout seems to be more stable in terms of its fluctuations during the middle of the training process, between epoch 100 and 150, however it still fails to match our method in the final loss as well as final accuracy. This highlights the problems laid out in the introduction section where a model loses on accuracy in an attempt to not overfit.

\subsection{RQ3: Intrinsic Regularization}

\begin{figure}[]
\centering
\subfloat[CIFAR10\label{fig:cifar50}]{%
  \includegraphics[width=0.33\textwidth]{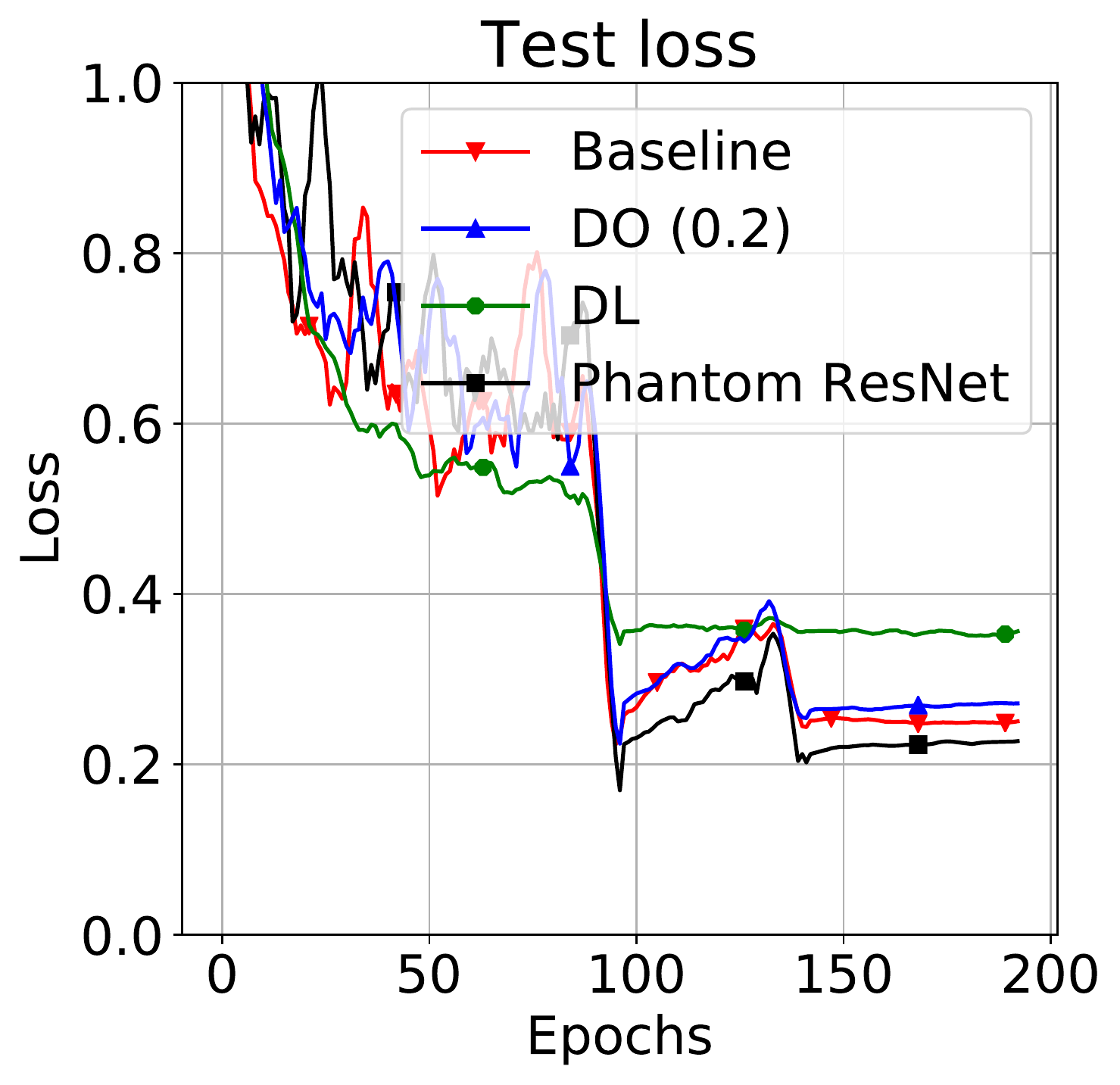}%
}\hfil
\subfloat[FashionMNIST\label{fig:fashion50}]{%
  \includegraphics[width=0.33\textwidth]{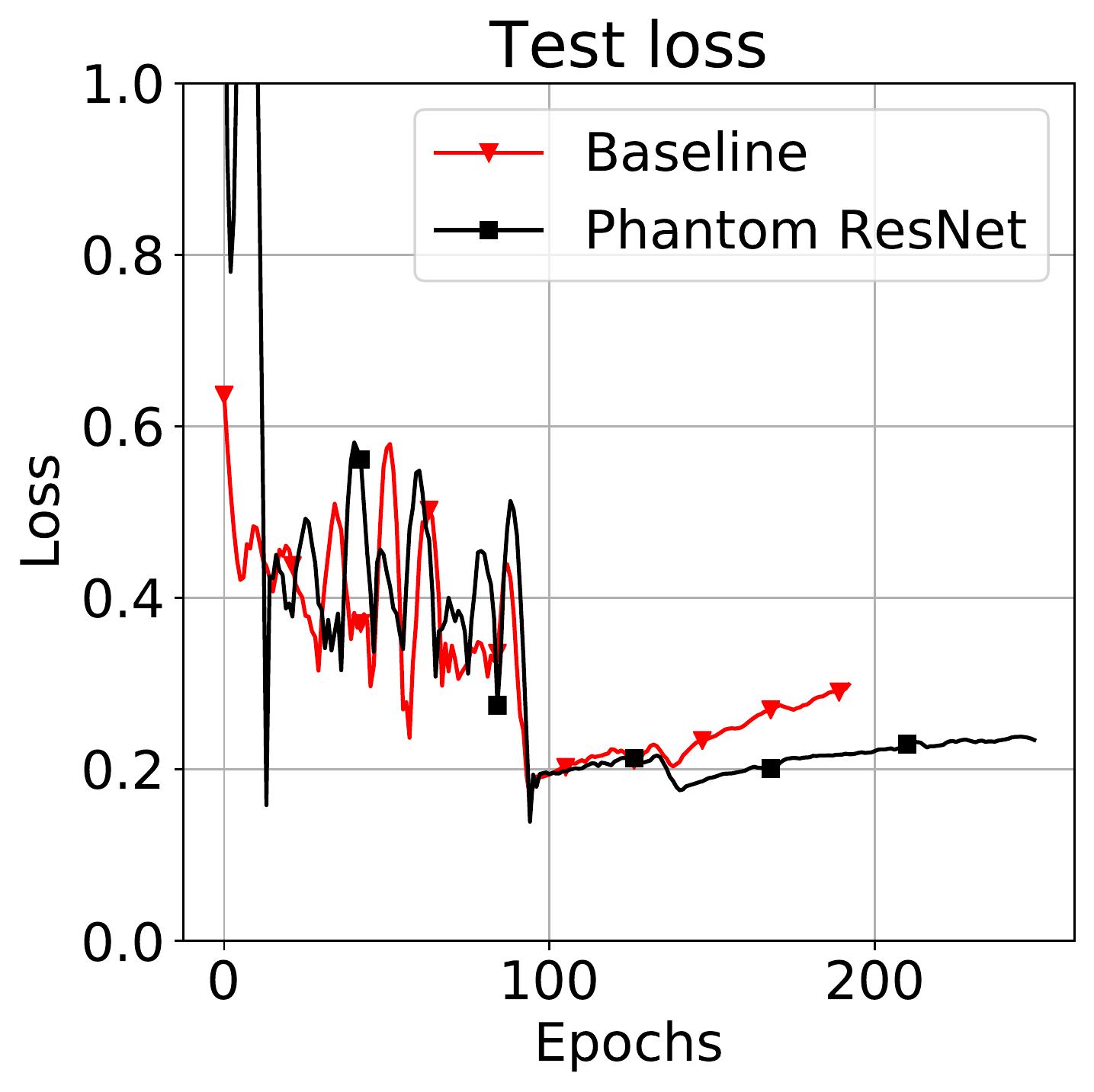}%
}\hfil
% \subfloat[Caption1\label{fig:subim2}]{%
%   \includegraphics[width=0.33\textwidth]{figures/resnet50_trainLoss.pdf}%
% }
\caption{Intrinsic Regularization in ResNet50: The phantom embeddings prevent overfitting even when the training regime is specifically aiming to overfit.}
\label{fig:resnet50_plots}
\end{figure}

As stated earlier, training deep models are hampered by the model memorizing the training data and then showing poor performance on the test data. This problem comes to the forefront when dealing with a truly deep model like \resnetvv which comes with 0.88M trainable parameters. Training such a model from scratch requires an immense amount of data or a clever regularization scheme. The scheme needs to be searched for over several runs and hyper-parameter configurations. This is a time-consuming and expensive procedure since training \resnetvv can take up to 7-11 hours on a modern GPU. 
Our proposed method allows for the data samples to contribute not just to the learning but to the regularization as well. By intrinsically learning the regularization with the help of similar images and generalizing the weights of our embedding layer with our proposed phantom embeddings we are able to regularize the model as it trains. This behavior is on display in Fig. \ref{fig:resnet50_plots} where it can be seen that our model is leading to a marked lower test loss while the baseline models struggle to match its performance. Given enough time (days) an ideal configuration for the baselines could be arrived to match the performance of our model however, our model provides it without the need for the extensive search required by the baselines. 

In Fig. \ref{fig:fashion50} we intentionally allowed the models to run past their convergence point to see how the baseline and our model handle such cases. It can be seen that the baselines runs off and starts to overfit, leading to an increasing test loss while our method shows a noticably better performance and maintains a lower test loss. 

\subsection{Abalation Study}

In order to showcase the effect of different numbers of samples from the same class ($K$) we varied  $K$ from 1 (baseline) to 7 and in Tab. \ref{tab:cifar_ablation}. It was seen that while increasing $K$ led to increasing performance over the baselines, the percentage gain vs memory required didn't justify the use of higher $K$. All the results reported have, therefore been  conducted with $K=2$.
\begin{table}[]
\centering
\caption{Abalation Study: Investigating the effect of increasing K on the classification accuracy.}

\begin{tabular}{|l|l|l|l|l|}
\hline 
\multicolumn{1}{|l|}{\textbf{}}             & \multicolumn{4}{c|}{\textbf{Accuracy}}	\\ \hline	

\hline
Model     & K=1   & K=2   & K=3   & K=4  \\ \hline
ResNet 18 & 93.5  & 94.91 & 94.01 & 93.9 \\ \hline
ResNet 34 & 93.65 & 94.58 & 94.3  & 94.2 \\ \hline
\end{tabular}
\label{tab:cifar_ablation}
\end{table}

%!TEX root = ../main.tex

\section{Conclusion}

In this paper, we have shown how embedding spaces can be directly used to regularize deeper neural networks by creating phantom embeddings around the true data points by aggregating the embeddings together and then optimizing the model with the phantom embedding as a co-target. We have shown how our method outperforms the baselines two famous and competitive datasets. Our method also introduces an intrinsic regularization which enables us to train deeper models without an extensive hyper-parameter search.

%
% ---- Bibliography ----
%
% BibTeX users should specify bibliography style 'splncs04'.
% References will then be sorted and formatted in the correct style.
%
\bibliographystyle{splncs04}
\bibliography{bibliography}

\begin{thebibliography}{10}
\providecommand{\url}[1]{\texttt{#1}}
\providecommand{\urlprefix}{URL }
\providecommand{\doi}[1]{https://doi.org/#1}

\bibitem{chapelle2001vicinal}
Chapelle, O., Weston, J., Bottou, L., Vapnik, V.: Vicinal risk minimization.
  In: Advances in neural information processing systems. pp. 416--422 (2001)

\bibitem{deng2009imagenet}
Deng, J., Dong, W., Socher, R., Li, L.J., Li, K., Fei-Fei, L.: Imagenet: A
  large-scale hierarchical image database. In: 2009 IEEE conference on computer
  vision and pattern recognition. pp. 248--255. Ieee (2009)

\bibitem{code_DL}
Farzaneh, A.: Disturblabel-pytorch (2019),
  \url{https://github.com/amirhfarzaneh/disturblabel-pytorch}

\bibitem{girshick2015fast}
Girshick, R.: Fast r-cnn. In: Proceedings of the IEEE international conference
  on computer vision. pp. 1440--1448 (2015)

\bibitem{goodfellow2013maxout}
Goodfellow, I., Warde-Farley, D., Mirza, M., Courville, A., Bengio, Y.: Maxout
  networks. In: International conference on machine learning. pp. 1319--1327
  (2013)

\bibitem{he2015delving}
He, K., Zhang, X., Ren, S., Sun, J.: Delving deep into rectifiers: Surpassing
  human-level performance on imagenet classification. In: Proceedings of the
  IEEE international conference on computer vision. pp. 1026--1034 (2015)

\bibitem{he2016deep}
He, K., Zhang, X., Ren, S., Sun, J.: Deep residual learning for image
  recognition. In: Proceedings of the IEEE conference on computer vision and
  pattern recognition. pp. 770--778 (2016)

\bibitem{ioffe2015batch}
Ioffe, S., Szegedy, C.: Batch normalization: Accelerating deep network training
  by reducing internal covariate shift. arXiv preprint arXiv:1502.03167  (2015)

\bibitem{krizhevsky2014cifar}
Krizhevsky, A., Nair, V., Hinton, G.: The cifar-10 dataset. online: http://www.
  cs. toronto. edu/kriz/cifar. html  \textbf{55} (2014)

\bibitem{krizhevsky2012imagenet}
Krizhevsky, A., Sutskever, I., Hinton, G.E.: Imagenet classification with deep
  convolutional neural networks. In: Advances in neural information processing
  systems. pp. 1097--1105 (2012)

\bibitem{krogh1992simple}
Krogh, A., Hertz, J.A.: A simple weight decay can improve generalization. In:
  Advances in neural information processing systems. pp. 950--957 (1992)

\bibitem{lecun1989backpropagation}
LeCun, Y., Boser, B., Denker, J.S., Henderson, D., Howard, R.E., Hubbard, W.,
  Jackel, L.D.: Backpropagation applied to handwritten zip code recognition.
  Neural computation  \textbf{1}(4),  541--551 (1989)

\bibitem{kli}
Li, K.: kuangliu/pytorch-cifar (2017),
  \url{https://github.com/kuangliu/pytorch-cifar}

\bibitem{nair2010rectified}
Nair, V., Hinton, G.E.: Rectified linear units improve restricted boltzmann
  machines. In: ICML (2010)

\bibitem{pratt1993discriminability}
Pratt, L.Y.: Discriminability-based transfer between neural networks. In:
  Advances in neural information processing systems. pp. 204--211 (1993)

\bibitem{simard1998transformation}
Simard, P.Y., LeCun, Y.A., Denker, J.S., Victorri, B.: Transformation
  invariance in pattern recognition—tangent distance and tangent propagation.
  In: Neural networks: tricks of the trade, pp. 239--274. Springer (1998)

\bibitem{simonyan2014very}
Simonyan, K., Zisserman, A.: Very deep convolutional networks for large-scale
  image recognition. arXiv preprint arXiv:1409.1556  (2014)

\bibitem{sirinukunwattana2016locality}
Sirinukunwattana, K., Raza, S.E.A., Tsang, Y.W., Snead, D.R., Cree, I.A.,
  Rajpoot, N.M.: Locality sensitive deep learning for detection and
  classification of nuclei in routine colon cancer histology images. IEEE
  transactions on medical imaging  \textbf{35}(5),  1196--1206 (2016)

\bibitem{srivastava2014dropout}
Srivastava, N., Hinton, G., Krizhevsky, A., Sutskever, I., Salakhutdinov, R.:
  Dropout: a simple way to prevent neural networks from overfitting. The
  journal of machine learning research  \textbf{15}(1),  1929--1958 (2014)

\bibitem{srivastava2015training}
Srivastava, R.K., Greff, K., Schmidhuber, J.: Training very deep networks. In:
  Advances in neural information processing systems. pp. 2377--2385 (2015)

\bibitem{szegedy2015going}
Szegedy, C., Liu, W., Jia, Y., Sermanet, P., Reed, S., Anguelov, D., Erhan, D.,
  Vanhoucke, V., Rabinovich, A.: Going deeper with convolutions. In:
  Proceedings of the IEEE conference on computer vision and pattern
  recognition. pp.~1--9 (2015)

\bibitem{vapnik1998statistical}
Vapnik, V., Vapnik, V.: Statistical learning theory wiley. New York  \textbf{1}
  (1998)

\bibitem{vapnik2015uniform}
Vapnik, V.N., Chervonenkis, A.Y.: On the uniform convergence of relative
  frequencies of events to their probabilities. In: Measures of complexity, pp.
  11--30. Springer (2015)

\bibitem{xiao2017fashion}
Xiao, H., Rasul, K., Vollgraf, R.: Fashion-mnist: a novel image dataset for
  benchmarking machine learning algorithms. arXiv preprint arXiv:1708.07747
  (2017)

\bibitem{xie2016disturblabel}
Xie, L., Wang, J., Wei, Z., Wang, M., Tian, Q.: Disturblabel: Regularizing cnn
  on the loss layer. In: Proceedings of the IEEE Conference on Computer Vision
  and Pattern Recognition. pp. 4753--4762 (2016)

\bibitem{zhang2017mixup}
Zhang, H., Cisse, M., Dauphin, Y.N., Lopez-Paz, D.: mixup: Beyond empirical
  risk minimization. arXiv preprint arXiv:1710.09412  (2017)

\end{thebibliography}

\end{document}